%% file: ijcai23.tex
\documentclass{article}
\usepackage{ijcai23}

\usepackage{times}
\usepackage{soul}
\usepackage{url}
\usepackage[hidelinks]{hyperref}
\usepackage[utf8]{inputenc}
\usepackage[small]{caption}
\usepackage{graphicx}
\usepackage{amsmath}
\usepackage{amsthm}
\usepackage{booktabs}
\usepackage{algorithm}
\usepackage{algorithmic}
\usepackage[switch]{lineno}

\usepackage{makecell} 
\usepackage{multirow}

\usepackage{mathtools}
\usepackage{subfigure}
\usepackage{amsfonts}
\usepackage[table]{xcolor}

\newtheorem{definition}{Definition}[section]

\newtheorem{lemma}{Lemma}[section]
\newtheorem{corollary}{Corollary}[section]

\newcommand{\citet}[1]{\citeauthor{#1}~\shortcite{#1}}

\title{FedSampling: A Better Sampling Strategy for Federated Learning}

\author{
Tao Qi$^1$
\and
Fangzhao Wu$^2$\thanks{The corresponding author.}\and
Lingjuan Lyu$^{3}$\and
Yongfeng Huang$^{1,4,5}$ \And
Xing Xie$^2$
\affiliations
$^1$Department of Electronic Engineering, Tsinghua University, Beijing 100084, China \\
$^2$Microsoft Research Asia, Beijing 100080, China \\
$^3$Sony AI, 1-7-1 Konan Minato-ku Tokyo 108-0075, Japan \\
$^4$Zhongguancun Laboratory, Beijing 100094, China  \\
$^5$ Institute for Precision Medicine of Tsinghua University, Beijing 102218, China
\emails
\{taoqi.qt,wufangzhao\}@gmail.com \and
Lingjuan.Lv@sony.com \and \\
yfhuang@tsinghua.edu.cn \and
xingx@microsoft.com
}

\begin{document}

\maketitle

\begin{abstract}
Federated learning (FL) is an important technique for learning models from decentralized data in a privacy-preserving way.
Existing FL methods usually uniformly sample clients for local model learning in each round.
However, different clients may have significantly different data sizes, and the clients with more data cannot have more opportunities to contribute to model training, which may lead to inferior performance.
In this paper, instead of client uniform sampling, we propose a novel data uniform sampling strategy for federated learning (\textit{FedSampling}), which can effectively improve the performance of federated learning especially when client data size distribution is highly imbalanced across clients.
In each federated learning round, local data on each client is randomly sampled for local model learning according to a probability based on the server desired sample size and the total sample size on all available clients.
Since the data size on each client is privacy-sensitive, we propose a privacy-preserving way to estimate the total sample size with a differential privacy guarantee. 
Experiments on four benchmark datasets show that \textit{FedSampling} can effectively improve the performance of federated learning.
\end{abstract}

\input{data/Introuduction}
\input{data/Relatework}
\input{data/Approach}

\input{data/Experiment}

\input{data/Conclusion}

\section*{Acknowledgments}
This work was supported by the National Key Research and Development Program of China 2022YFC3302104, Tsinghua University Initiative Scientific Research Program of Precision Medicine under Grant number 2022ZLA0073, Tsinghua-Toyota Joint Research Fund 20213930033, and China Postdoctoral Science Fundation (No.2022M721892).

\bibliographystyle{named}
\bibliography{ijcai23}

\end{document}

%% file: data/Introuduction.tex
\section{Introduction}

\begin{figure}
    \centering
    \resizebox{0.49\textwidth}{!}{
    \includegraphics{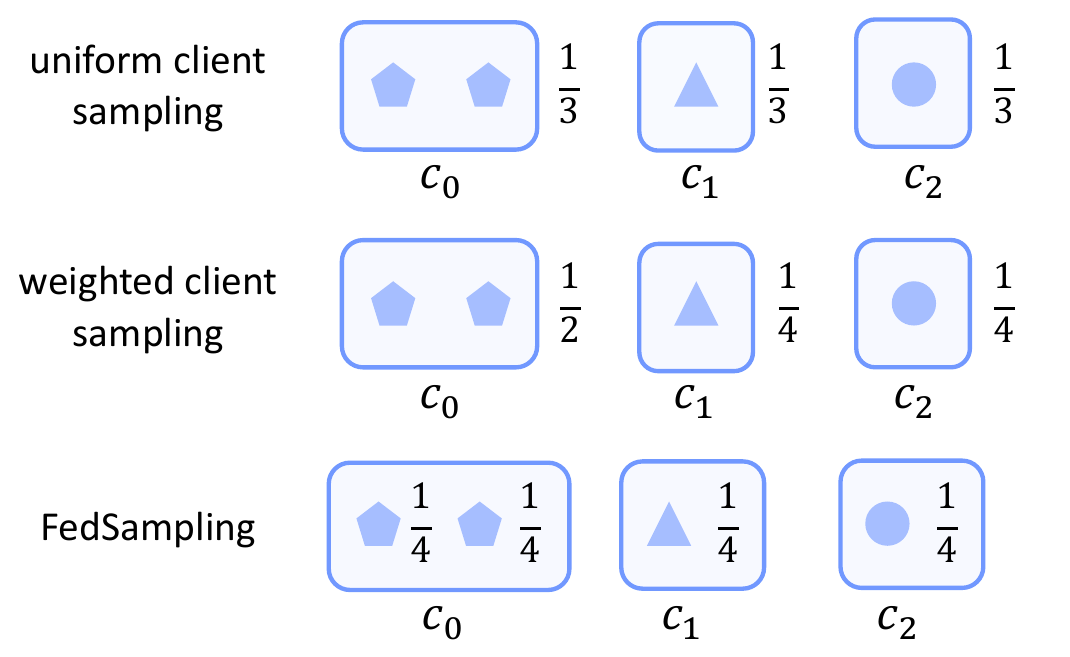}
    }
    \caption{Toy example of different sampling methods, where the numbers indicate sampling probabilities. Both uniform and weighted client sampling are biased in data exploitation.}
    \label{fig.motivation}
\end{figure}

Federated learning aims to utilize decentralized data to train machine learning models~\cite{mcmahan2017communication}, and has become a popular privacy-preserving machine learning paradigm~\cite{dayan2021federated,bai2021advancing,fedspeech,rothchild2020fetchsgd}.
The mainstream federated learning framework~\cite{reddi2020adaptive,yang2018applied,ldpfl} usually uniformly samples some clients for local model training in each round.
However, sizes of samples on different clients may largely differ~\cite{fraboni2021impact}.
The uniform client sampling may prevent the effective exploitation of the clients with more data for model training and lead to a suboptimal performance~\cite{fraboni2021impact}.
For instance, consider a scenario with $3$ clients (Fig.~\ref{fig.motivation}), in which the client $c_0$, $c_1$ and $c_2$ keep $2$, $1$ and $1$ samples respectively, and we need to select a client for training in each round.
For uniform client sampling methods, although the sampling probability of each client is uniform and identical to $\frac{1}{3}$, the weights of samples in different clients for model updating are biased.
For the client $c_0$, two samples in it will be used for training at the same time and thus their weights are identical to $\frac{1}{2}$, while the weights of samples in the client $c_1$ and $c_2$ are identical to $1$.
Thus, the model training may pay more attention to samples in the client $c_1$ and $c_2$, which leads to a biased data exploitation.

To address this issue, some recent methods have enabled the server to track the local sample size in each client to perform weighted client sampling~\cite{li2019convergence,wang2020tackling,fraboni2021clustered}.
In these methods, the sampling probability of a client is the ratio of its local sample size to the sample size of all clients.
However, these methods usually over-emphasize clients with more data, thus fail to satisfy an unbiased exploitation of samples, leading to sub-optimal performance.
In the example of Fig.~\ref{fig.motivation}, for weighted client sampling methods, the sampling probabilities of data in the client $c_0$ are identical to $\frac{1}{2}$ while the sampling probabilities of data in the client $c_1$ and $c_2$ are identical to $\frac{1}{4}$.
Thus, half of the total samples in the client $c_0$ will be sampled for training more frequently than the remaining half of the total samples, which is still biased in data exploitation.
Besides, in some certain scenarios (e.g., financial transactions), the local sample size represents the frequency of some privacy-sensitive behaviors.
Thus, allowing the server to track local sample sizes may also arouse privacy concerns~\cite{gerber2018explaining}.

In short, existing client-level sampling methods are usually biased in the data exploitation due to the dependent or non-identical data sampling.
In this paper, we propose a uniform data sampling framework (named \textit{FedSampling}) to improve the effectiveness of federated learning.
At each training round of \textit{FedSampling}, each sample in each client is independently sampled with an identical probability, where the probability is the ratio of the server desired sample size to the total sample size in all available clients.
Considering the privacy of local sample size in each client, we propose a privacy-preserving method to estimate the total sample size based on the local differential privacy (LDP) technique.
Each client randomly chooses to send the server a true or a randomly-generated local sample size to protect personal privacy, and the server can estimate an unbiased total sample size from the randomized responses.
Besides, the analysis on the utility and privacy of \textit{FedSampling} is also provided.
We conduct extensive experiments on four benchmark datasets across different domains.
Results show that \textit{FedSampling} can outperform many recent FL methods, especially under imbalanced data size distribution and non-IID data distribution, and meanwhile achieve an effective trade-off between utility and privacy.

%% file: data/Relatework.tex
\section{Related Work}

Due to the importance of user privacy, privacy-persevering machine learning techniques have attracted increasing attention~\cite{avdiukhin2021federated,achituve2021personalized,jumper2021highly}.
Federated learning can utilize massive private data distributed on local clients to benefit the training of a shared ML model~\cite{hamer2020fedboost,fedfairavg,li2020federated}, and thereby has become a popular privacy-aware machine learning framework~\cite{dayan2021federated,yoon2021federated,huang2021fl,bai2021advancing}.
Existing methods usually follow a similar paradigm, where a server uniformly samples some clients and collects their local model parameters to update a global model, iteratively~\cite{liang2021fedrec,yuan2021federated,fraboni2021clustered}.
For example, \citet{mcmahan2017communication} proposed to locally train models on clients for multiple steps, and further average local model parameters from sampled clients to update the global model.
\citet{reddi2020adaptive} proposed a federated version of the Adam optimization algorithm to smooth the model training, where the current local parameters are combined with historical local parameters to learn the global model.
However, the sample size of different clients may be highly imbalanced, and clients with more data cannot be sampled more frequently to contribute more to the model training in these methods, which usually results in a sub-optimal performance.

Some methods explored weighted client sampling to handle this problem~\cite{li2019convergence,wang2020tackling,fraboni2021clustered}.
For example, \citet{li2019convergence} proposed to first sample clients for local model training based on the ratios between their local samples and the total samples in all clients, then average local model updates in the server to update the global model.
However, client-level weighting usually over-emphasizes the clients with more data and also cannot achieve a uniform data exploitation.
Besides, these methods need the server to track the local sample size of each client, which may be impracticable in some scenarios due to privacy concerns.
Different from these methods, we propose a data-level uniform sampling strategy, which can achieve a similar data exploitation like centralized learning to improve the performance of federated model training.

%% file: data/Approach.tex
\begin{figure*}[t]
    \centering
     \resizebox{0.98\textwidth}{!}{
     \includegraphics{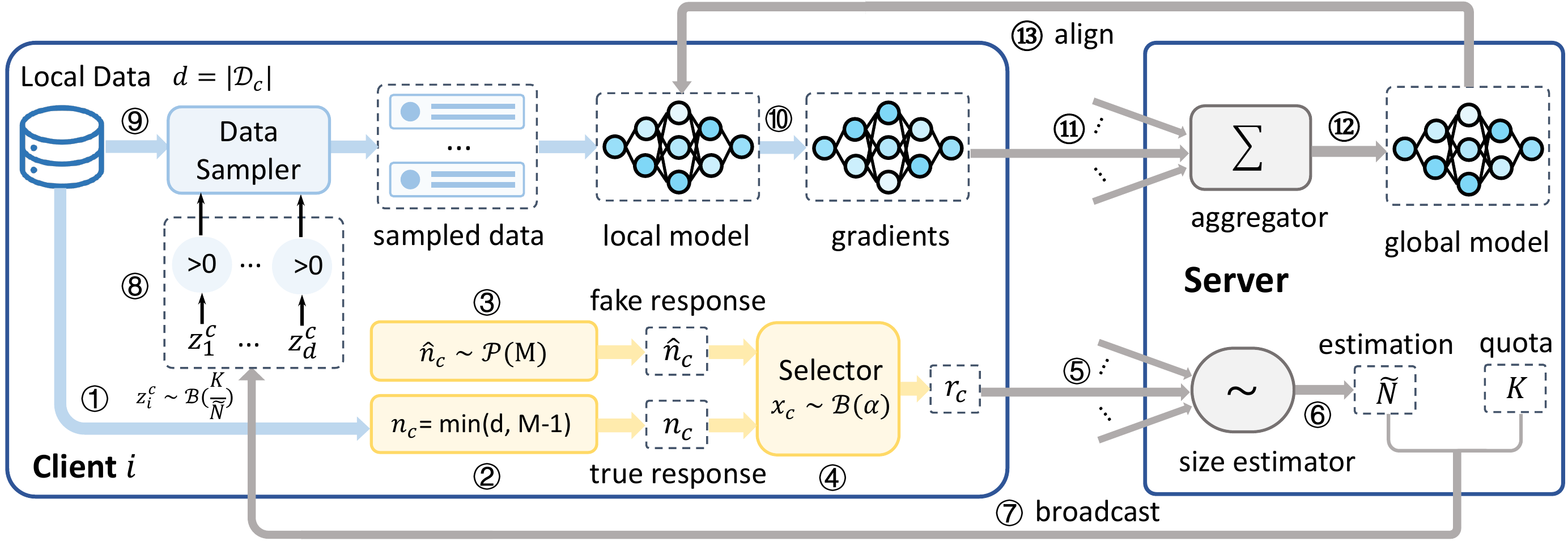}
     }
    \caption{The framework of \textit{FedSampling}.}
    \label{fig.framework}
\end{figure*}

\section{FedSampling}

\subsection{Problem Formulation}

In our work, we assume that there are $H$ available clients that can participate in the federated learning, and the set of available clients is denoted as $\mathcal{C}$.
Each client $c$ locally keeps a private dataset $\mathcal{D}_c$ and never shares it with the outside, where $\mathcal{D}_c = \{s_i|i=1,2,...,|\mathcal{D}_c|\}$ and $s_i$ denotes the $i$-th local sample.
Moreover, the local sample size $|\mathcal{D}_c|$ of each client $c$ is also assumed to be privacy-sensitive and cannot be disclosed.
Besides, there is a server that takes charge of maintaining an ML model and scheduling the workflow of the federated model training. 
The core problem of this work is how to effectively sample decentralized data for model training to improve the effectiveness of federated learning.

\subsection{Uniform Data Sampling}

Uniform data sampling has the potential to improve the effectiveness of federated learning since it has a similar data exploitation pattern like centralized learning.
Intuitively, by allowing the server to track the local sample size of each client, we can allocate quotas for each client according to the size of its local samples to achieve a uniform data sampling.
However, in many scenarios (e.g., medicine and financial transactions) the size of local data is privacy-sensitive~\cite{gerber2018explaining}, and thereby cannot be tracked by the server.
For instance, the size of medical records in a client represents the frequency of medical activities, which can be highly privacy-sensitive for many users.
In our work, we propose a unified method that can uniformly and independently sample decentralized data for federated training without the collection of local sample sizes.

Without loss of generalization, we assume the server needs $K$ samples to collaboratively update the model in a training round.
Besides, we assume the server can obtain an estimated number $\widetilde{N}$ of total samples in all available clients\footnote{We will introduce how we obtain $\widetilde{N}$ in the next section.}.
In \textit{FedSampling}, the server first broadcasts $K$ and $\widetilde{N}$ to each client and then each client can locally sample local data for training according to its local sample size and the server desire.
Take a client $c$ as example, we first independently draw a random variable $z^c_i$ from a Bernoulli distribution $\mathcal{B}(\frac{K}{\widetilde{N}})$ for each local sample in $\mathcal{D}_c$, where the ratio $\frac{K}{\widetilde{N}}$ is the probability of assigning $1$ to $z^c_i$, which means the $i$-th local sample $s^c_i$ will be sampled for local training.
In this way, each sample in each client can be independently sampled for training with identical probabilities.
In addition, the expected number of samples for participating in this training round is $N\mathbb{E}[\frac{K}{\widetilde{N}}]$, which can asymptotically converge to $K$ and meet the server demand (the convergence is discussed in the following section).
After the client $c$ obtains a set $\mathcal{S}_c$ of locally sampled data, the client $c$ will employ the current model $\Theta$ to calculate model gradients $\textbf{g}_s$ on each sample $s\in \mathcal{S}_c$, and build the local model update based on normalized local gradients: $\textbf{G}_c = \frac{1}{K} \sum_{s\in\mathcal{S}_c} \textbf{g}_s$.
Then the local update is uploaded to the server for the global model updating.

\subsection{Privacy-Preserving Ratio Estimation}

Next, we will introduce how to estimate the ratio of the server desired sample number $K$ to the total sample number $N$ in a privacy-preserving way.
Intuitively, we may bypass this problem if we sample data based on a fixed ratio $r$ (e.g., $1\%$) instead of a total sample number-aware ratio $\frac{K}{N}$.
However, this will arise new challenges that need to be carefully addressed.
Specifically, in this naive method, the local update $\textbf{G}_c$ in the client $c$ cannot be locally normalized by the number of sampled data in this round due to the lack of knowledge of it (i.e., $r\times N$).
The client $c$ can only obtain a local model update like mainstream FL frameworks by averaging local gradients: $\textbf{G}_c = \frac{1}{|\mathcal{S}_c|}\sum_{s\in\mathcal{S}_c} \textbf{g}_s$.
On one hand, to obtain an unbiased global model updating, the server in this naive method needs to employ the number of locally sampled data in each client to weight the aggregation of the corresponding local updates.
However, this will cause privacy leakage due to the disclosure of the number of locally sampled data in each client.
On the other hand, the naive method can uniformly average the local model updates to avoid the potential privacy leakage.
However, this will lead to a biased model update and result in a sub-optimal performance.
Thus, the naive method mentioned above is not optimal for uniform data sampling.

Next, we will introduce a carefully designed method to estimate this ratio in a privacy-preserving way.
Specifically, the server first queries clients for their local sample sizes.
When a client $c$ receives the query, it will first generate a true response $n_c$ by clipping the local sample size $|\mathcal{D}_c|$ via a size threshold $M$: $n_c = \min(|\mathcal{D}_c|,M-1)$.
Then the client $c$ will generate a fake response by drawing a random variable $\hat{n}_c$ from a uniform multinomial distribution: $\hat{n}_c\sim \mathcal{P}(M)$, where we randomly select an integer from $1$ to $M-1$ with identical probabilities and assign it to $\hat{n}_c$.
Furthermore, the client $c$ will draw a random variable $x_c$ from a Bernoulli distribution for the response selection:
\begin{equation}
\label{eq.response}
    r_c = x_cn_c + (1-x_c)\hat{n}_c, \quad x_c\sim \mathcal{B}(\alpha),
\end{equation}
where $\alpha$ and $1-\alpha$ denotes the probability of assigning $1$ and $0$ to $x_c$ respectively, and $r_c$ is the selected response.
Under the protection of this method, it is difficult for the server to obtain the accurate knowledge on the local sample size of any client.
The privacy protection ability of our privacy-preserving ratio estimation method can satisfy the $\epsilon$-LDP, which will be discussed in the following section.
Furthermore, after receiving responses from clients, the server further aggregates them to estimate the total sample size $N$:
\begin{equation}
    \widetilde{N} =(R - \frac{(1-\alpha)M|\mathcal{C}|}{2} )/\alpha, \quad R = \sum_{c\in\mathcal{C}} r_c,
    \label{eq.total}
\end{equation}
where $\widetilde{N}$ is the estimated total sample size and is proved to be unbiased in the following section.
Then the server distributes the ratio of $K$ to $\widetilde{N}$ to clients for uniform data sampling.


\begin{algorithm}
	\caption{Workflow of \textit{FedSampling}}
  \label{algo}
	\begin{algorithmic}[1]
		\FORALL{$t\leftarrow 1$ to $T$}
		    \STATE Broadcast model $\Theta_{t-1}$ and query sample sizes
    		 \FOR{$c$ in $\mathcal{C}$}
    		    \STATE Draw $x_c$ from $\mathcal{B}(\alpha)$, $\hat{n}_c$ from
    		    $\mathcal{P}(M)$
    		    \STATE Obtain $r_c$ via Eq.~\ref{eq.response}
    		    \STATE Send $r_c$ to the server
    		\ENDFOR
    		\STATE Calculate $\widetilde{N}$ via Eq.~\ref{eq.total}, and broadcast $\widetilde{N}$, $K$
    		\FOR{$c$ in $\mathcal{C}$}
                \STATE Initialize an empty data queue $\mathcal{S}_c$
    		    \FOR{$i\leftarrow 1$ to $|\mathcal{D}_c|$} 
    		            \STATE Draw $z^c_i \sim \mathcal{B}(\frac{K}{\widetilde{N}})$
    		            \STATE Push $s^c_i$ into $\mathcal{S}_c$ if $z^c_i>0$
    		    \ENDFOR
    		    \STATE Build local model update based on samples in $\mathcal{S}_c$
    		    \STATE Send local model update to the server
    	    \ENDFOR
    	    \STATE $\Theta_t = \Theta_{t-1}+\eta\sum_{\textbf{G}\in\mathcal{G}}\textbf{G}$
	    \ENDFOR
	\end{algorithmic}
\end{algorithm}

\subsection{Federated Training Framework}
Next, we will introduce the workflow of federated training in \textit{FedSampling}, which is summarized in Fig.~\ref{fig.framework} and Algorithm~\ref{algo}.
The workflow of \textit{FedSampling} is similar to the mainstream FL methods in local model training and global model updating, while different in data sampling.
The $t$-th training round in \textit{FedSampling} includes the following steps.

First, the server broadcasts the current model $\Theta_{t-1}$ for model aligning and queries each client for the local sample size, where $\Theta_{t-1}$ is the model in the $t-1$-th round.
Then clients respond to the server based on Eq.~\ref{eq.response} and the server estimates the total sample size $\widetilde{N}$ via Eq.~\ref{eq.total}.
Second, the server broadcasts $K$ and $\widetilde{N}$ to clients. 
Then clients can further uniformly sample data to obtain the local updates and upload them to the server.
Third, until receiving all local updates, the server aggregates them to update the global model: $\Theta_t = \Theta_{t-1}+\eta\sum_{\textbf{G}\in\mathcal{G}}\textbf{G}$, where $\mathcal{G}$ is the set of received local updates, and $\eta$ is the learning rate.

\subsection{Discussions on Utility and Privacy}

Next, we will analyze the utility and privacy of \textit{FedSampling} in theory and compare it with mainstream FL methods.

\subsubsection{Utility Analysis on FedSampling}

\begin{definition}
An FL algorithm is \textbf{data-level unbiased}, iff any data can be independently sampled for a training step with identical probabilities to centralized training.
\end{definition}
Based on this definition, data-level unbiased FL methods have a similar data exploitation pattern with centralized learning.
Thus, data-level unbiased FL methods have the potential to more effectively exploit decentralized data under imbalanced data size distribution and even the non-IID data distribution.
We remark that FL methods based on the uniform or weighted client sampling are not data-level unbiased, due to the violation of independent or identical sampling conditions.
Furthermore, based on Lemma~\ref{lemma.error}, in \textit{FedSampling} data can be sampled with probabilities identical to centralized learning in a training round.
Besides, since \textit{FedSampling} controls the sampling of data via independent random variables, and thereby it is data-level unbiased.

\input{data/proof1}

\subsubsection{Privacy Analysis on FedSampling}

For privacy protection, most FL methods disclose sample sizes and may violate privacy constrains.
We are the first to propose a privacy-preserving ratio estimation method that can protect privacy in the local sample sizes.
Our method can achieve $\epsilon$-LDP in a training round via Lemma~\ref{theo.ldp}.
\begin{definition}
A randomized mechanism $\mathcal{M}$($\cdot$) satisfies $\epsilon$-LDP, iff for two arbitrary input $x$ and $x'$, and any output $y\in range(\mathcal{M})$, the following inequation holds:
\begin{equation}
    Pr[\mathcal{M}(x)=y]\leq e^{\epsilon}\cdot Pr[\mathcal{M}(x')=y].
\end{equation}
The privacy budget $\epsilon$ quantifies the privacy protection ability, where a smaller $\epsilon$ means better privacy protection.
\end{definition}
\begin{lemma}\label{theo.ldp}
Given an arbitrary size threshold $M$, \textit{FedSampling} can meet $\epsilon$-LDP in protecting the privacy of local sample size in each client, when $\alpha = \frac{e^\epsilon-1}{e^\epsilon+M-2}$.
\end{lemma}
\input{data/proof2}

We can further de-link responses from the same client at different iterations via client anonymity~\cite{ldpfl} to avoid the accumulation of privacy budget.
Thus, \textit{FedSampling} can meet $\epsilon$-LDP in the whole training process.
Besides, the disclose of local updates may also leak user privacy.
We remark that this concern is out of the scope of this paper, as there are already many works~\cite{lyu2020privacy} that protect local updates via LDP. It is straightforward to apply many of them to \textit{FedSampling} for further privacy protection.

%% file: data/proof1.tex
\begin{lemma}
\label{lemma.error}
Let $p(x)$ and $\hat{p}(x)$ denote the probability of a sample $x$ that can participate in a training step in the centralized learning and \textit{FedSampling}. 
The mean square error between $p(\cdot)$ and $\hat{p}(\cdot)$ asymptotically converges to $0$.
\begin{equation}
 \lim_{|\mathcal{C}|\to \infty} \mathbb{E}[(\hat{p}(x)-p(x))^2] = 0.
\end{equation}
\end{lemma}


\begin{proof}
Denote the data size distribution as $\mathcal{P}_{\mathcal{D}}$, then we can view the selected response $r_c$ as a combination of three random variables (Eq.~\ref{eq.combination}), and further obtain its expectation:
\begin{equation}
 r_c = x_c n_c + (1-x_c)\hat{n}_c, \quad n_c\sim \mathcal{P}_{\mathcal{D}}, \label{eq.combination}
\end{equation}
\begin{equation}
\mathbb{E}[r_c] = \alpha \overline{n} + (1-\alpha)\frac{M}{2},
\end{equation}
where $\overline{n}$ is the averaged sample size of different clients.
Furthermore, we prove $\widetilde{N}$ is an unbiased estimation of $N$: 
\begin{equation} 
    \mathbb{E}[R] = \mathbb{E}[\sum_{c\in\mathcal{C}} r_c] = \alpha N +(1-\alpha)\frac{M|\mathcal{C}|}{2},
\end{equation}
\begin{equation}
\mathbb{E}[\widetilde{N}] = (\mathbb{E}[R] - \frac{(1-\alpha)M|\mathcal{C}|}{2})/\alpha = N. \label{eq.exp}
\end{equation}
Furthermore, without the loss of generalization, we assume that there are $K$ samples selected for model training in a single step.
Based on the uniform data sampling strategy in the centralized learning, and the sampling strategy in \textit{FedSampling}, we can obtain the following equations:
\begin{equation}
   {p}(x) = \frac{K}{N},\quad  \hat{p}(x) = \frac{K}{\widetilde{N}}. 
\end{equation}
Let $f(x) = (\frac{1}{x}-\frac{1}{N})^2$, based on the Taylor series, we have:
\begin{equation}
    f(x) = f(\mathbb{E}[x]) + \sum_{l=1}\frac{f^{l}(\mathbb{E}[x]) }{l!}(x-\mathbb{E}[x])^l.
\end{equation}
Let $x = \widetilde{N}$, and note that $f(N)=0$ and $f^{l}(N) = \frac{(-1)^l(l-1)l!}{N^{l+2}}$, then we have the following equation:
\begin{equation}
   \mathbb{E}[f(\widetilde{N})] = \sum_{l=2}\frac{(-1)^l(l-1)}{N^{l+2}} \mathbb{E}[(\widetilde{N}-N)^l]. \label{eq.taylor}
\end{equation}
Furthermore, the following inequation holds:
\begin{equation}
   \mathbb{E}[(\frac{1}{\widetilde{N}}-\frac{1}{N})^2] <\sum_{l=2}\frac{(l-1)}{N^{l+2}} \mathbb{E}[|\widetilde{N}-\mathbb{E}[\widetilde{N}]|^l].
\end{equation}
Further, based on tight version of the $C_r$ inequation~\cite{ineq}, we have:
\begin{equation}
     \mathbb{E}[|\widetilde{N}-\mathbb{E}[\widetilde{N}]|^l]<\frac{2|\mathcal{C}|-1}{\alpha^l}\mathbb{E}[|r_c-\mathbb{E}[r_c]|^l].
\end{equation}
Note that $\mathbb{E}[|r_c-\mathbb{E}[r_c]|^l]$ is a constant determined by the data size distribution and the setting of our size anonymization method (i.e., the value of $\alpha$ and $M$), and we denote it as $\delta_l$.
Note that $|\mathcal{C}|<N$, then we obtain this inequation:
\begin{equation}
 \mathbb{E}[(\hat{p}(x)-p(x))^2]< K^2\sum_{l=2}\frac{2(l-1)}{|\mathcal{C}|^{l+1}\alpha^l} \delta_l.
\end{equation} 
Note that $\delta_l$, and $\alpha$ are constants and irrelevant with $|\mathcal{C}|$, then we can obtain our expected conclusion:
\begin{equation}
     \lim_{|\mathcal{C}|\to \infty}  \mathbb{E}[(\hat{p}(x)-p(x))^2] <  K^2\lim_{|\mathcal{C}|\to \infty}  \sum_{l=2}\frac{2(l-1)}{|\mathcal{C}|^{l+1}\alpha^l} \delta_l = 0.
\end{equation}
\end{proof}

\begin{corollary}
The mean square error between $\frac{NK}{\widetilde{N}}$ and $K$ can asymptotically converge to $0$:
\begin{equation}
     \lim_{|\mathcal{C}|\to \infty}  \mathbb{E}[(\frac{KN}{\widetilde{N}}- K)^2] = 0.
\end{equation}
\end{corollary}

%% file: data/proof2.tex
\begin{proof}
Let $\mathcal{M}(\cdot)$ denote the privacy preserving ratio estimation method.
Consider an arbitrary client $c$ with a true response $n_c$, and a protected response $r_c$, then we obtain:
\begin{equation}
\label{eq.ldp.large}
    Pr[\mathcal{M}(n_c) = r_c | r_c = n_c] = \alpha + \frac{1-\alpha}{M-1},
\end{equation}
\begin{equation}
\label{eq.ldp.small}
    Pr[\mathcal{M}(n_c) = r_c | r_c \not= n_c] = \frac{1-\alpha}{M-1},
\end{equation}
where we can find the Eq.~\ref{eq.ldp.large} always larger or equal than Eq.~\ref{eq.ldp.small}.
Furthermore, for two arbitrary $c$ and $c'$ in $\mathcal{C}$ and any output $y\in \{y|y=1,2,...,M-1\}$, we have:
\begin{equation}
\begin{split}
    & \max_{c,c',y} \frac{Pr[\mathcal{M}(n_c)=y]}{Pr[\mathcal{M}(n_{c'})=y]}\\
& = \frac{Pr[\mathcal{M}(n_c)=y|y=n_c]}{Pr[\mathcal{M}(n_{c'})=y|y\not=n_{c'}]} \\
& = \frac{(M-2)\alpha+1}{1-\alpha}.
\end{split}
\end{equation}
Furthermore, according to the definition of $\epsilon$-LDP, we have:
\begin{equation}
    e^{\epsilon} =  \max_{c,c',y} \frac{Pr[\mathcal{M}(r_c)=y]}{Pr[\mathcal{M}(r_{c'})=y]} = \frac{(M-2)\alpha+1}{1-\alpha}.
\end{equation}
Thus, when $M$ and $\epsilon$ is fixed, we can obtain:
\begin{equation}
    \alpha = \frac{e^\epsilon-1}{e^\epsilon + M - 2}.
\end{equation}
\end{proof}

%% file: data/Experiment.tex
\section{Experiment}

\subsection{Datasets and Experimental Settings}
\label{sec.setting}

\input{Table/ExpTable1}

Experiments are conducted on four benchmark datasets: a text dataset \textit{MIND}~\cite{wu2020mind}, two Amazon review datasets (i.e., \textit{Toys} and \textit{Beauty})~\cite{mcauley2015image}, and an image dataset \textit{EMNIST}~\cite{cohen2017emnist}.
First, \textit{MIND}, \textit{Toys} and \textit{Beauty} are used to evaluate model effectiveness under imbalanced data size distribution.
These three datasets are used for the text classification task, where \textit{Text-CNN}~\cite{kimcnn} and the mainstream NLP model \textit{Transformer}~\cite{vaswani2017attention} are used as the basic model.
Data in \textit{Toys} and \textit{Beauty} can be naturally partitioned into clients based on user IDs, which follows a long-tail distribution (shown in Appendix).
Besides, since texts in \textit{MIND} dot not include user information, we partition \textit{MIND} based on a typical long-tail distribution, i.e., log-normal distribution.
The average size of samples in each client is set to $2$ and we adjust the variance $\sigma$ of the distribution to control the data size imbalance.
$\sigma$ is set to $4$ in experiments.
Second, we use \textit{MIND} and \textit{EMNIST} to verify how non-IID data distribution affects different methods.
Motivated by \citet{mcmahan2017communication}, we sort training data in each dataset by their labels, and partition them into different clients of size $300$ and $6000$ respectively.
Thus, distributions of local data in different clients are varied.
We use the \textit{ResNet} network~\cite{he2016deep} for the image classification on \textit{EMNIST}.
Marco-F1 and Accuracy are used for classification task evaluation.
In our \textit{FedSampling} method, the size threshold $M$ and privacy budget $\epsilon$ are set to $300$ and $3$ respectively.
The learning rate $\eta$ is set to $0.05$, and the number $K$ of samples for participating in a training round is set to 2048.
All hyper-parameters are selected on the validation set.
More model details are in the Appendix and Codes (\url{https://github.com/taoqi98/FedSampling}).

\subsection{Performance Evaluation}
\label{sec.exp.evaluation}

We compare \textit{FedSampling} with several recent federated learning methods.
We first compare the standard \textit{FedAvg} algorithm and three adaptive federated optimization methods proposed to improve model effectiveness under heterogeneous data distribution, including:
(1) \textit{FedAvg}~\cite{mcmahan2017communication}: uniformly selecting clients for federated model training, and averaging local model updates to train the global model.
(2) \textit{FedYogi}~\cite{reddi2020adaptive}: a federated version of the \textit{Yogi} optimization algorithm that smooths global model updates based on momentum strategy~\cite{reddi2018adaptive}.
(3) \textit{FedAdagrad}~\cite{reddi2020adaptive}, a federated version of the \textit{Adagrad} optimization algorithm~\cite{lydia2019adagrad}.
(4) \textit{FedAdam}~\cite{reddi2020adaptive}, a federated version of the \textit{Adam} optimization algorithm~\cite{kingma2014adam}.
We remark all that these baseline FL methods are based on uniform client sampling according to their original settings.
We also apply the centralized model training (\textit{Centralization}) as a baseline method to quantify the performance decline of different federated learning methods.

We repeat each experiment five times and show average performance and standard deviations in Table~\ref{table.text}.
First, compared with \textit{Centralization}, baseline FL methods have serious performance declines on all datasets.
This is because the size distributions of many real-world user data are usually imbalanced.
However, baseline FL methods usually uniformly select some clients to participate in a training round, and thereby fail to effectively exploit massive training data in some long-tail clients.
Second, our \textit{FedSampling} method can consistently achieve comparable performance with \textit{Centralization}.
This is because our \textit{FedSampling} method has a sample-level sampling strategy, which can uniformly and independently sample data to achieve a uniform and unbiased data exploitation like centralized learning.

\begin{figure}
    \centering
    \resizebox{0.47\textwidth}{!}{
    \includegraphics{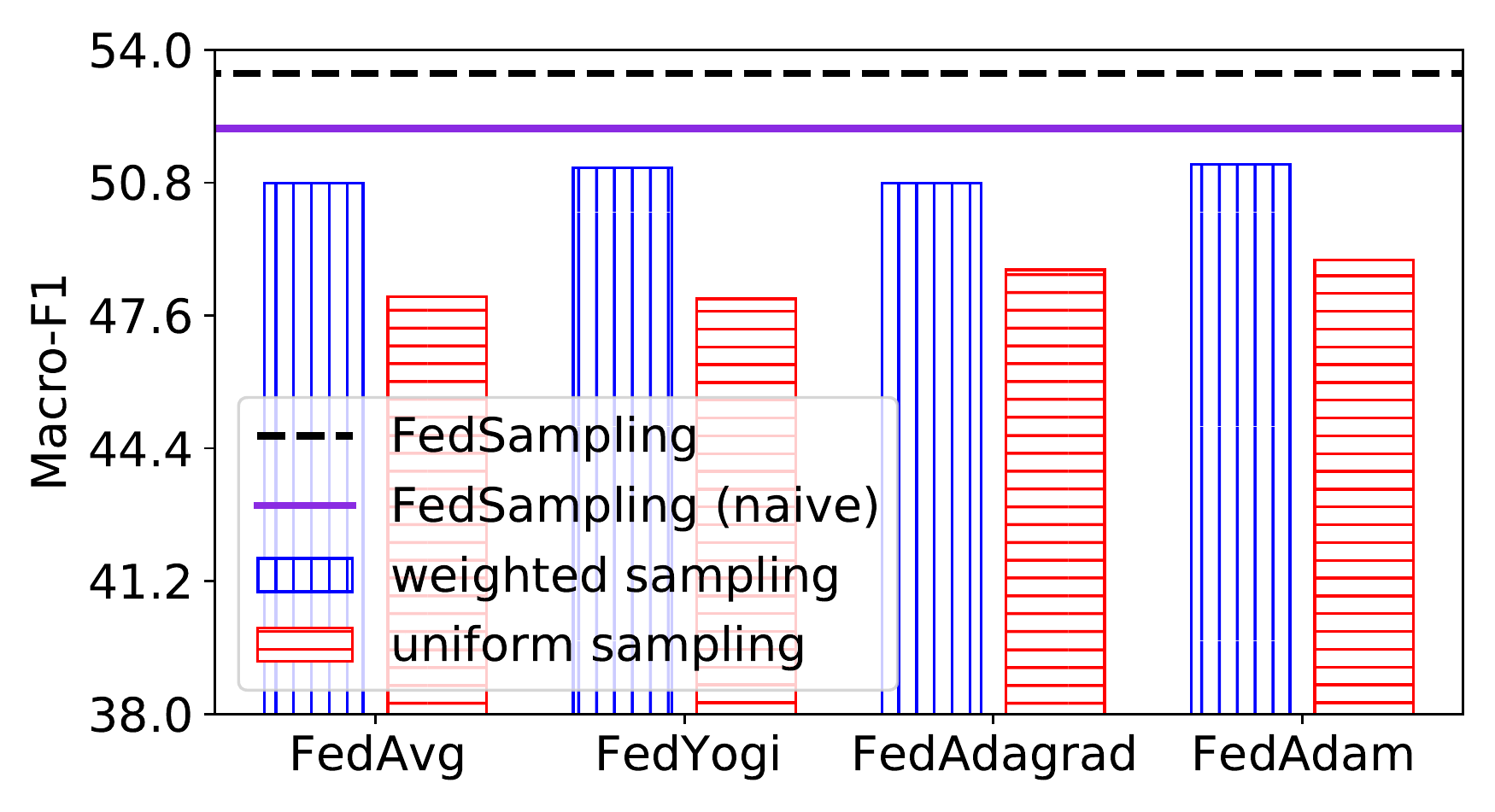}
    }
    \caption{Comparisons of \textit{FedSampling} with weighted client sampling on different federated learning methods. }

    \label{fig.sampling}
\end{figure}

To more effectively exploit decentralized data for model training, some existing federated learning methods~\cite{li2019convergence,wang2020tackling} sample clients based on local sample sizes.
Thus, we apply the weighted client sampling strategy to baseline FL methods for further comparisons.
Due to space limitations, we only show results on \textit{MIND} with the \textit{Transformer} model in the following sections.
Results are presented in Fig.~\ref{fig.sampling}, and we have several findings.
First, we can find that weighted client sampling significantly improves the performance of baseline FL methods.
This is because based on weighted client sampling long-tail clients have more opportunities to participate in a training round, and thereby massive data in them can be more effectively exploited.
Second, compared with \textit{Centralization} and \textit{FedSampling}, performance of baseline FL methods with weighted sampling are still inferior.
This is because weighted client sampling usually emphasizes long-tail clients during model training while head clients can also accumulate non-neglect samples.
These methods still cannot achieve uniform data exploitation, which leads to sub-optimal performance.
Different from these methods, we propose a sample-level sampling strategy which can independently and uniformly sample decentralized data for training.
In addition, the weighted client sampling strategy requires the server to track local samples sizes of each client, which may arouse privacy concerns due to the privacy sensitivity of local sample sizes.
Different from them, \textit{FedSampling} protects the local sample size of each client via a privacy-preserving ratio estimation method with a theoretical privacy guarantee.
Besides, we also compare the naive method introduced in approach section with \textit{FedSampling}.
Results show that performance of the naive method is inferior to \textit{FedSampling}.
This is because to avoid the privacy leakage on sample sizes, the naive method needs to uniformly average local updates to learn the global model, which is biased in data exploitation.

\begin{figure}
    \centering
    \resizebox{0.47\textwidth}{!}{
    \includegraphics{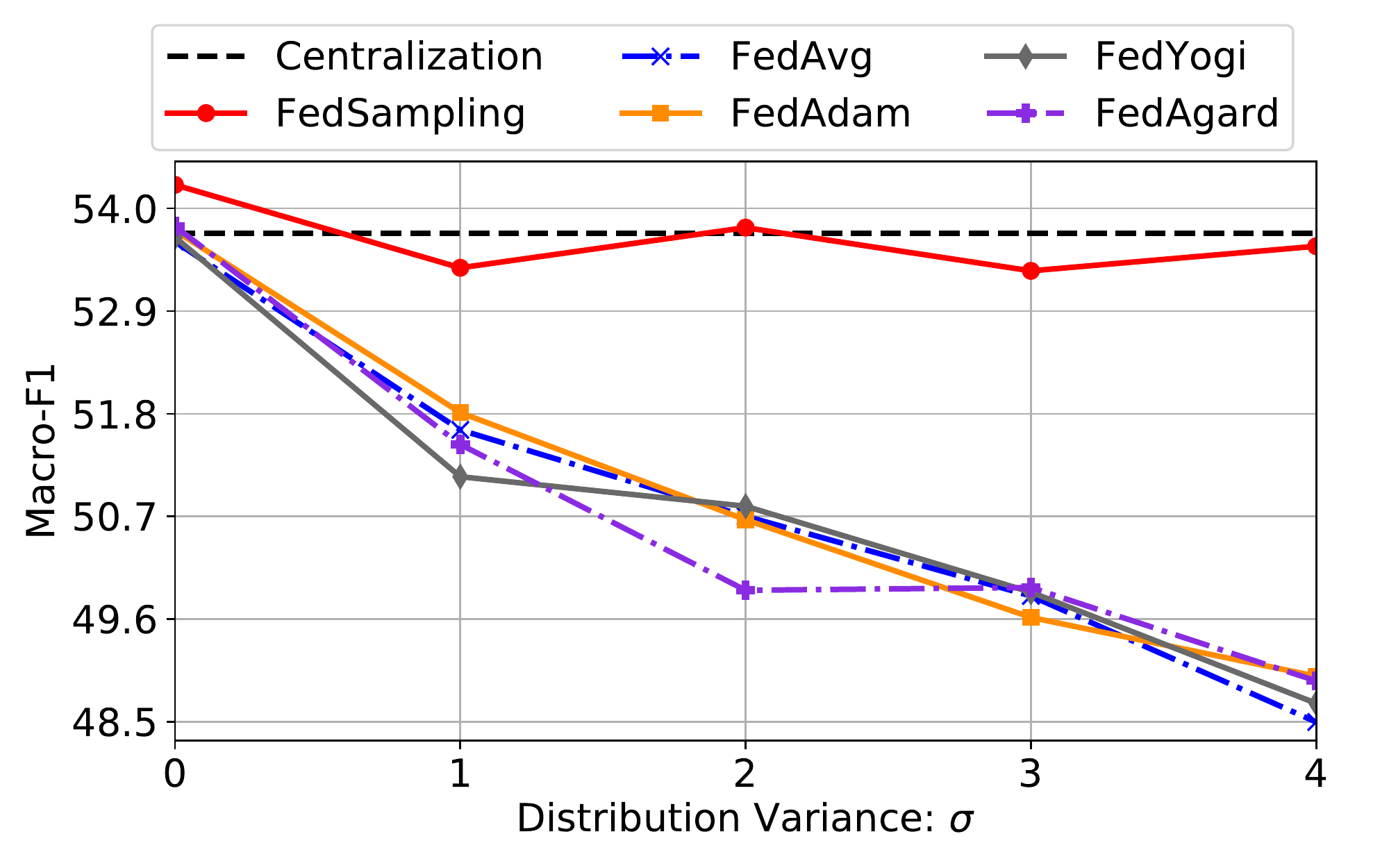}
    }
    \caption{Influence of data size imbalance degree.}
    \label{fig.distribution}
\end{figure}

\subsection{Influence of Data Size Distribution Imbalance}

\label{sec.exp.size}

Next, we will analyze how the imbalance of data size distribution affects different methods.
According to our experimental settings, the data partition of \textit{MIND} is based on a log-normal distribution.
Thus, we can control the size imbalance by adjusting the variance $\sigma$ of the log-normal distribution, where larger $\sigma$ leads to more imbalanced sample sizes.
Results in Fig.~\ref{fig.distribution} show the influence of $\sigma$ on the effectiveness of different methods, from which can we summarize two major phenomenons.
First, with the increase of $\sigma$, the performance of baseline methods consistently declines.
This is because the increase of $\sigma$ makes sample sizes more imbalanced, where long-tail clients can keep more data.
Thus, there will be more data that fails to be effectively exploited by baseline FL methods.
Second, \textit{FedSampling} has comparable performance with \textit{Centralization} under different $\sigma$.
This is because \textit{FedSampling} achieves uniform sampling in data level and can effectively learn models under imbalanced data size distribution.

\subsection{Comparisons under Non-IID Distribution}
\label{sec.exp.noniid}

\begin{figure}
    \centering
    \resizebox{0.47\textwidth}{!}{
    \includegraphics{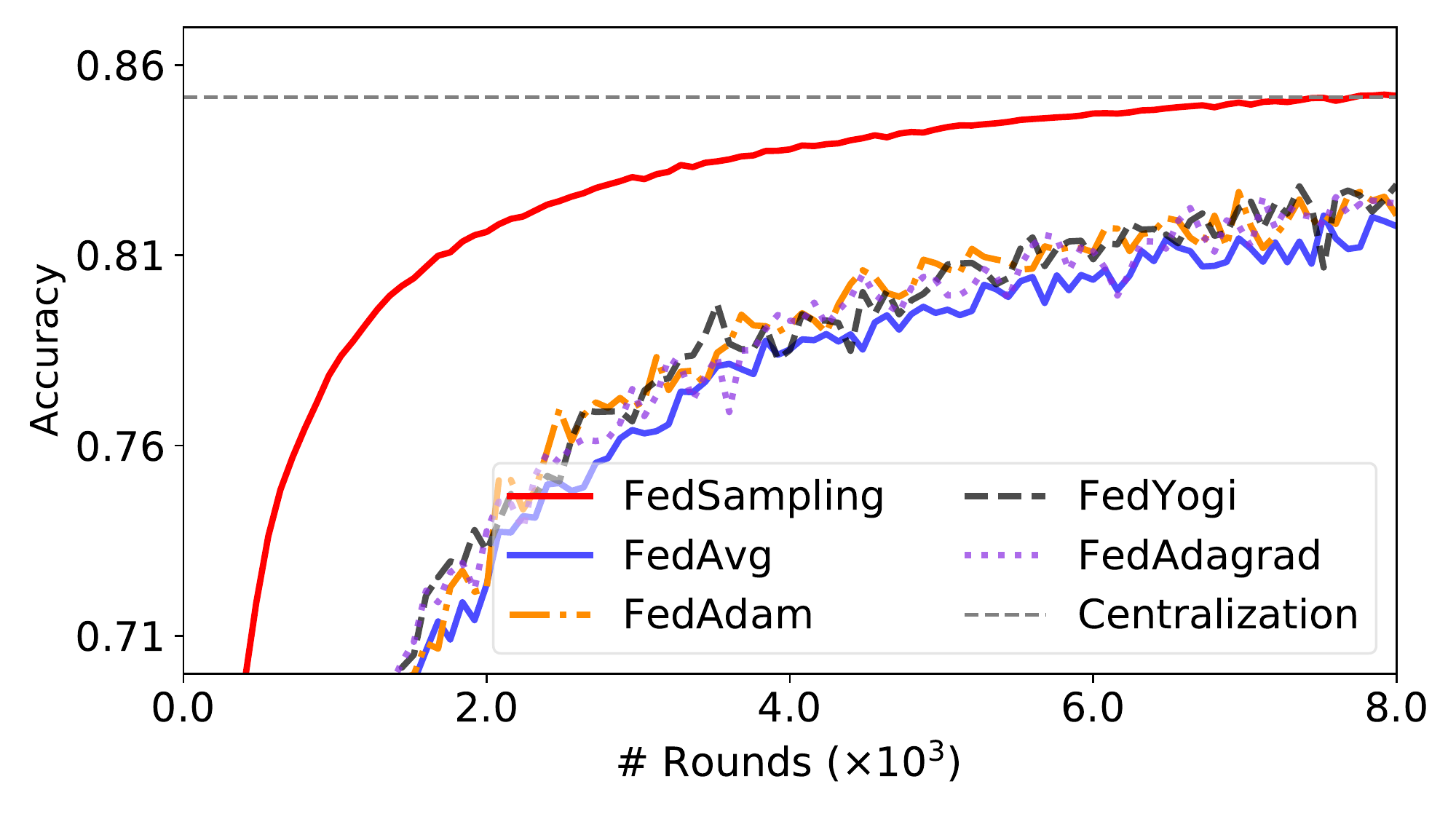}}

    \caption{Performance comparisons under non-IID data distribution on the \textit{EMNIST} dataset. }
    \label{fig.noniid}
\end{figure}

Since \textit{FedSampling} can uniformly sample decentralized data, it has the potential to improve the effectiveness of FL under non-IID data distribution.
Next, we compare different methods under the non-IID data distribution.
We set clients to have equal sample size, i.e., $300$, which makes no differences for baseline FL methods to perform uniform or weighted client sampling.
Thus, in this section, we only show results of baseline FL methods with uniform client sampling.
Results are shown in Fig.~\ref{fig.noniid}, from which we can find the convergence of \textit{FedSampling} is faster and smoother.
This is because distributions of data on different clients are highly heterogeneous, where most clients only keep data with the same classification category.
However, baseline FL methods sample data for training at the client level, which results in heterogeneous local model updates and hurts the model convergence.
Different from these methods, \textit{FedSampling} can independently select samples in different clients for model training with identical opportunities.
Thus, \textit{FedSampling} can achieve a similar data sampling pattern with the canonical uniform data sampling in centralized learning, and improve the effectiveness of FL models under the non-IID data distribution.

\subsection{Utility and Privacy Analysis}

\label{sec.exp.tradeoff}

\begin{figure}
    \centering
    \resizebox{0.47\textwidth}{!}{
    \includegraphics{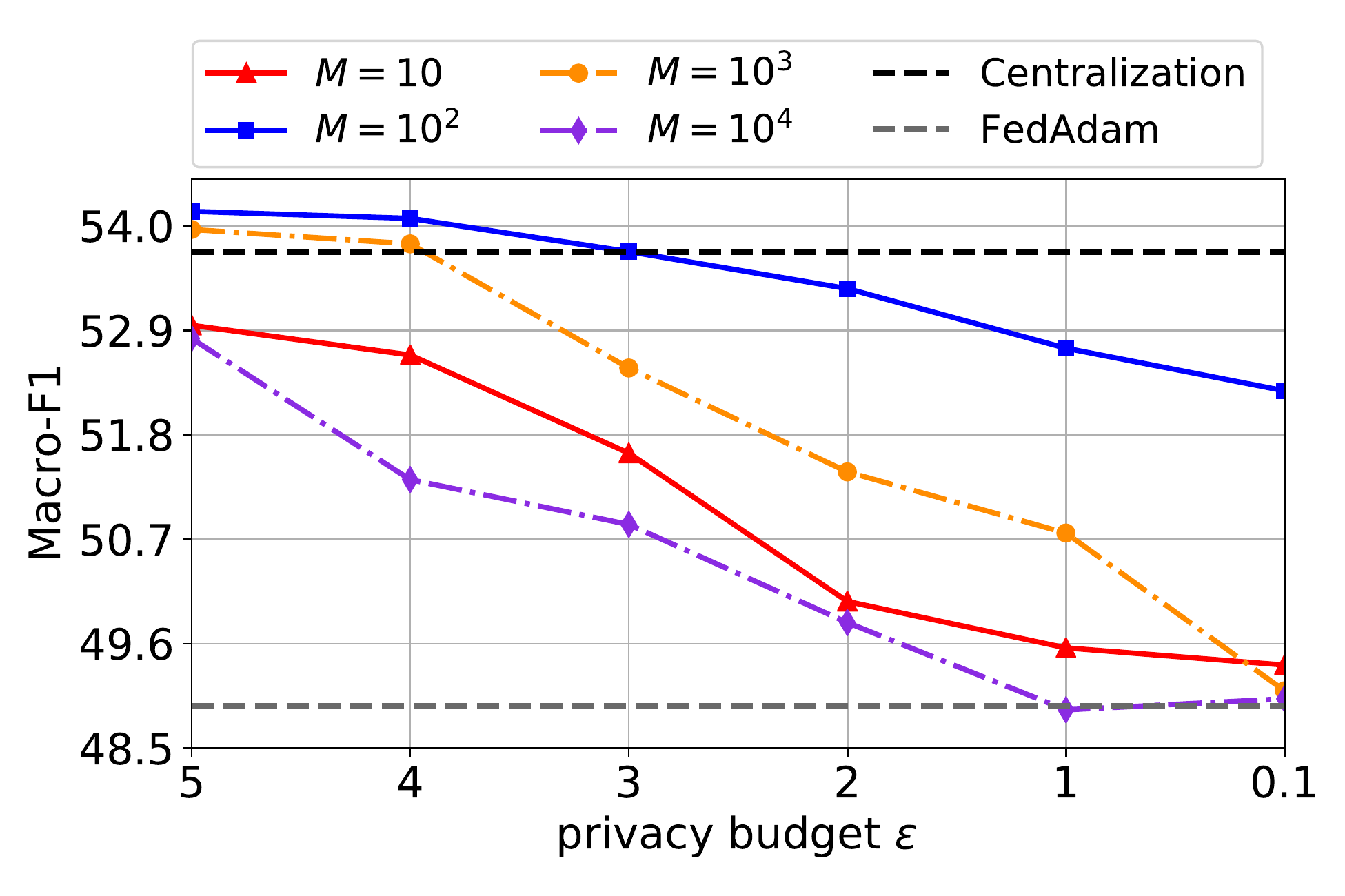}
    }
    \caption{Utility and privacy analysis of \textit{FedSampling}. }
    \label{fig.hyper}
\end{figure}

Next, we will analyze the trade-off between utility and privacy of our \textit{FedSampling} method.
As shown in Fig.~\ref{fig.hyper}, given a size threshold $M$, we evaluate the performance of \textit{FedSampling} under different privacy budget $\epsilon$.
First, with a constant size threshold $M$, a smaller privacy budget $\epsilon$ usually leads to a worse model performance.
This is because a smaller privacy budget $\epsilon$ increases the probability of clients for sending a fake response to the server, which hurts the accuracy of the total sample number estimation and the quality of data sampling.
Second, when the privacy budget  $\epsilon$ is fixed, \textit{FedSampling} can achieve the best performance under a moderate size threshold $M$, e.g., $10^2$.
This is because when $M$ is too small, many true responses may be truncated by a small value and cause bias on the estimation of total sample size.
Besides, according to Lemma~\ref{theo.ldp}, when the privacy budget $\epsilon$ is fixed, a larger size threshold $M$ requires larger probabilities of sending fake responses, which hurts the accuracy of data size distribution estimation.
Thus, we select a moderate value of size threshold $M$ (i.e., 100) and privacy budget $\epsilon$ (i.e., $3$) in experiments.

%% file: Table/ExpTable1.tex
\begin{table*}[h]

\centering
\resizebox{0.98\textwidth}{!}{
\begin{tabular}{cc|cc|cc|cc}
\Xhline{1.5pt}
\multirow{2}{*}{Model}       & \multirow{2}{*}{Training Algorithm} & \multicolumn{2}{c|}{MIND}       & \multicolumn{2}{c|}{Toys} & \multicolumn{2}{c}{Beauty}  \\ \cline{3-8} 
                             &                                     & Macro-F1       & Accuracy       & Macro-F1       & Accuracy       & Macro-F1       & Accuracy       \\ \hline
\multirow{6}{*}{Text-CNN} & Centralization     & {51.52}$\pm$0.57&{71.14}$\pm$0.45 & {39.61}$\pm$1.13&{63.71}$\pm$0.22 & {43.90}$\pm$0.97&{62.20}$\pm$0.67\\ 
& FedAvg             & 48.11$\pm$0.66&69.23$\pm$0.73 & 35.32$\pm$0.78&61.63$\pm$0.33 & 38.44$\pm$1.43&60.75$\pm$0.36\\
& FedYogi            & 49.12$\pm$0.71&68.92$\pm$0.40 & 35.62$\pm$2.34&61.22$\pm$0.39 & 38.77$\pm$0.89&60.35$\pm$0.91\\
& FedAdagrad         & 48.55$\pm$0.92&67.74$\pm$1.89 & 34.69$\pm$0.70&60.63$\pm$1.36 & 37.20$\pm$1.90&60.64$\pm$0.70\\
& FedAdam            & 48.54$\pm$0.65&68.22$\pm$0.50 & 35.27$\pm$1.59&61.35$\pm$0.32 & 39.09$\pm$0.80&60.43$\pm$1.05\\
& FedSampling          & \textbf{51.33}$\pm$0.62&\textbf{71.15}$\pm$0.30 & \textbf{40.15}$\pm$1.27&\textbf{63.41}$\pm$0.74 & \textbf{43.04}$\pm$0.83&\textbf{62.96}$\pm$0.16\\ \Xhline{1.pt}
\multirow{6}{*}{Transformer} & Centralization     & {53.73}$\pm$0.62&{72.19}$\pm$0.28 & {41.86}$\pm$0.96&{63.56}$\pm$0.57 & {44.31}$\pm$0.70&{62.92}$\pm$0.48\\ 
& FedAvg             & 50.46$\pm$0.99 & 70.74$\pm$0.52 & 38.68$\pm$0.93&60.30$\pm$2.06 & 37.82$\pm$1.36&60.41$\pm$0.27\\
& FedYogi            & 50.94$\pm$0.59 & 70.29$\pm$0.53 & 37.75$\pm$1.87&61.44$\pm$0.36 & 38.10$\pm$1.07&60.17$\pm$0.33\\
& FedAdagrad        & 50.99$\pm$0.68 & 70.65$\pm$0.48 & 38.06$\pm$0.61&59.69$\pm$1.60 & 38.59$\pm$1.56&59.87$\pm$0.51\\
& FedAdam            & 50.69$\pm$0.58 & 70.83$\pm$0.28 & 37.58$\pm$0.77&60.59$\pm$1.24 & 38.44$\pm$1.42&60.65$\pm$0.46\\
& FedSampling          & \textbf{53.43}$\pm$0.57&\textbf{71.98}$\pm$0.37 & \textbf{41.63}$\pm$1.12&\textbf{64.03}$\pm$0.46 & \textbf{43.47}$\pm$0.94&\textbf{62.67}$\pm$0.60 \\ \Xhline{1.5pt}
\end{tabular}
}
\caption{Results on the text classification task. Best results in federated settings are in bold.}
\label{table.text}
\end{table*}


%% file: data/Conclusion.tex
\section{Conclusion}

In this paper, we propose a uniform data sampling strategy for federated learning (named \textit{FedSampling}), which can achieve an uniform data exploitation.
In each round of \textit{FedSampling}, each sample in each client is independently sampled for model training according to an identically probability, which is based on the ratio of the server desired sample size to the total sample size in all available clients.
We also propose a privacy-preserving method to estimate an unbiased total sample size with an LDP privacy guarantee on the local sample size of each client.
We conduct extensive experiments on four benchmark datasets.
Experimental results show that \textit{FedSampling} can outperform many FL methods especially under imbalanced data size distribution and non-IID data distribution.
Although effective, \textit{FedSampling} may increase the communication cost of federated learning.
In our future work we will explore to handle this problem.